\documentclass{article}

\usepackage[preprint,nonatbib]{confstyle}

\usepackage[utf8]{inputenc} 
\usepackage[T1]{fontenc}    
\usepackage{hyperref}       
\usepackage{url}            
\usepackage{booktabs}       
\usepackage{amsfonts}       
\usepackage{nicefrac}       
\usepackage{microtype}      
\usepackage{xcolor}         
\usepackage{graphicx}  
\usepackage{setspace}
\usepackage{multirow}
\usepackage{hyperref}

\title{HabitAction: A Video Dataset for Human Habitual Behavior Recognition}

\author{Hongwu Li,\ \ Zhenliang Zhang$^{*}$,\ \ Wei Wang$^{*}$\thanks{$^{*}$Corresponding authors. Hongwu Li is with the State Key Laboratory of Robotics and System, Harbin Institute of Technology, Harbin 150080, China.
E-mail: lhwhit@126.com.
Wei Wang and Zhenliang Zhang are with the State Key Laboratory of General Artificial Intelligence, Beijing Institute for General Artificial Intelligence (BIGAI), Beijing 100080, China.
E-mails: wangwei@bigai.ai, zlzhang@bigai.ai.
This work was supported by the National Natural Science Foundation of China (Grant No. 61976214). }
}

\makeatletter
\def\thanks#1{\protected@xdef\@thanks{\@thanks
        \protect\footnotetext{#1}}}
\makeatother

\begin{document}

\maketitle

\begin{abstract}
Human Action Recognition (HAR) is a very crucial task in computer vision. It helps to carry out a series of downstream tasks, like understanding human behaviors. Due to the complexity of human behaviors, many highly valuable behaviors are not yet encompassed within the available datasets for HAR, e.g., human habitual behaviors (HHBs). HHBs hold significant importance for analyzing a person's personality, habits, and psychological changes. To solve these problems, in this work, we build a novel video dataset to demonstrate various HHBs. These behaviors in the proposed dataset are able to reflect internal mental states and specific emotions of the characters, e.g., crossing arms suggests to shield oneself from perceived threats. The dataset contains 30 categories of habitual behaviors including more than 300,000 frames and 6,899 action instances. Since these behaviors usually appear at small local parts of human action videos, it is difficult for existing action recognition methods to handle these local features. Therefore, we also propose a two-stream model using both human skeletons and RGB appearances. Experimental results demonstrate that our proposed method has much better performance in action recognition than the existing methods on the proposed dataset. 
All the dataset and codes are available at \href{http://sites.google.com/view/hhb30}{\it{sites.google.com/view/hhb30}}.
\end{abstract}

\section{Introduction}
Human action recognition (HAR), which aims to predict action classes from videos, holds significant importance in the field of computer vision \cite{sun2022human}. 
Extensively studied for decades, HAR remains highly popular due to its vast range of potential applications including video surveillance \cite{khan2020human}, human-computer interaction \cite{rodomagoulakis2016multimodal}, and sports analysis \cite{soomro2015action}. Particularly, with the exponential growth of video data on the Internet in recent years, the need for HAR has become even more pronounced \cite{ravi2016deep}. 

Human behaviors are very diverse, and a large amount of video datasets have been proposed for HAR \cite{ionescu2013human3,soomro2012ucf101,wishart2022hmdb}. 
The existing HAR datasets mainly focus on the following types of actions: (1) Basic daily activities, such as walking, running, standing, sitting, etc. (2) Sports activities, such as basketball shooting, soccer kicking, swimming, etc. (3) Poses and gestures, like waving and thumbs-up. (4) Interactive behaviors, like handshaking and making phone calls. (5) Entertainment activities, like dancing and singing. 

Human habitual behavior (HHB) is a crucial but seldomly explored action category in HAR, which refers to the actions that individuals exhibit in certain contexts as a result of habits and without conscious awareness \cite{wood2016psychology}. 
These actions are often considered as manifestations of underlying psychological and habitual mappings in human cognition \cite{wood2002habits,neal2012habits}.
Understanding these HHBs holds significant importance for human belief and intent recognition, and can provide insights into the underlying psychological dynamics of individuals.
Therefore, the recognition of human habitual behaviors in videos has crucial implications for various applications, including emotion recognition, personalized recommendation systems, and intelligent human-computer interaction. 

The recognition and study of HHBs require large-scale and diverse datasets that cover a wide range of habitual behaviors and contexts. 
However, the scarcity of datasets specifically targeting habitual actions limits exploration and progress in this field.
Compared to other action recognition tasks, collecting data for habitual actions is more challenging \cite{roggen2010collecting,hermsen2016using}. 
Habitual actions typically occur naturally and unconsciously, making it difficult to capture them through traditional laboratory settings.
In addition, since these behaviors occur naturally and unobtrusively without drawing attention, they rarely appear in the titles of videos, making them difficult to retrieve directly.
Furthermore, due to the similarity of many of these behaviors, annotating these actions is also very challenging.

\begin{figure*}
    \centering
    \includegraphics[width=0.99\linewidth]{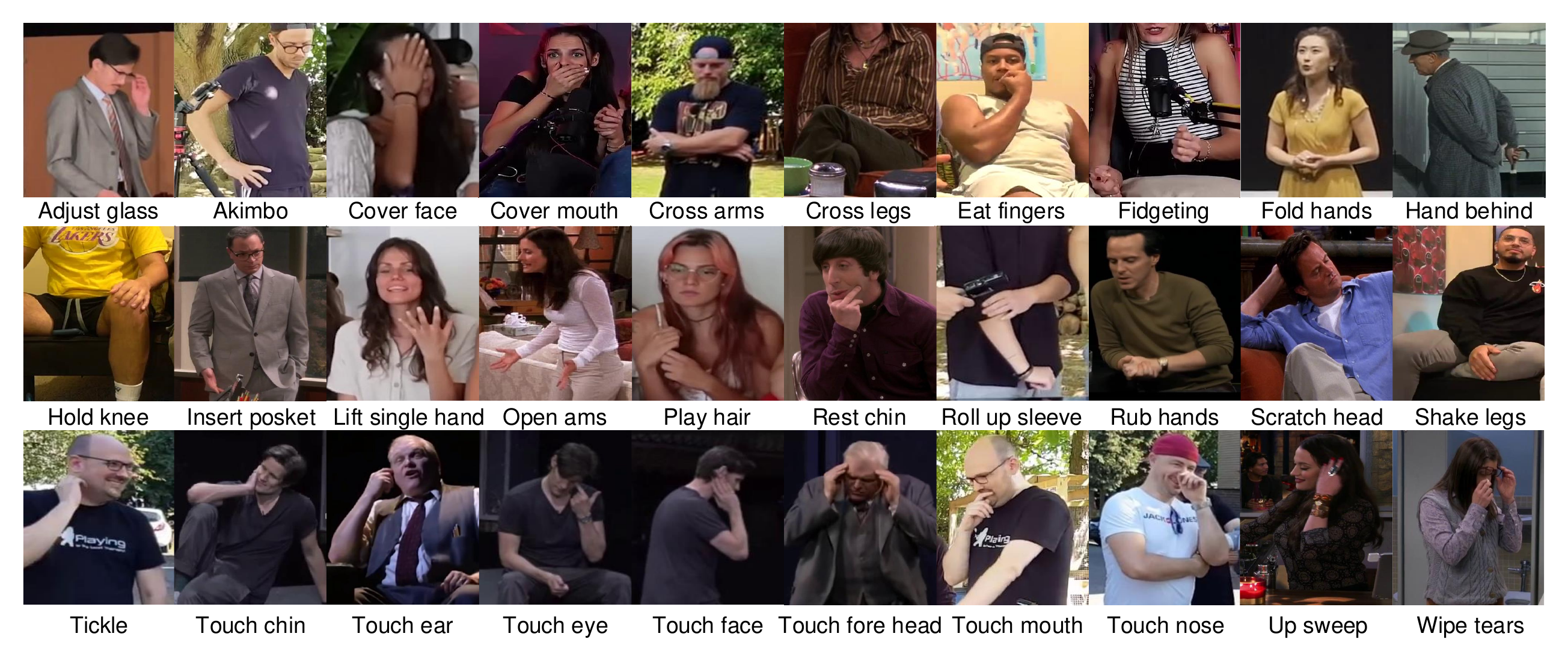}
    \vspace{-12pt}
    \caption{30 categories of human habitual behaviors collected in the proposed dataset. These behaviors generally occur unconsciously and reflect human internal mental states and emotions. Understanding and recognizing habitual behaviors are seldomly explored in previous work.}
   
    \label{fig:hhb}
\end{figure*}

In this work, we construct a new fine-grained HAR dataset, which contains a large number of human habitual behaviors.
To the authors' knowledge, there is no dataset built specifically for this type of action up to now.

The proposed dataset contains 30 categories of habitual actions (as shown in Fig. \ref{fig:hhb}) including more than 300,000 frames and 6,899 action instances.
Considering that these actions usually appear at small local parts of the video images, it is difficult for the existing action recognition methods to extract these local features.
We further propose a two-stream recognition model by using both human skeletons and RGB appearances. Extensive experimental results demonstrate that our model achieves much better performance than the existing methods in terms of recognition accuracy.

The main contributions of this work are summarized as follows:
\begin{itemize}
\item[$\bullet$] We propose a new task focusing on human habitual behaviors (HHBs), which reflects human personality traits and psychological changes. This work will promote advances in many fields, such as human-computer interaction, human-like action generation, etc.

\item[$\bullet$] We build a large dataset for fine-grained HHBs recognition. 
There is no such dataset built specifically for this kind
of actions before.

\item[$\bullet$] We propose a two-stream habitual behaviors recognition model, which is more suitable for subtle actions recognition and outperforms several state-of-the-art methods.
\end{itemize}

\section{Related Work}
\subsection{Human Action Recognition Datasets}
In recent years, there has been a notable rise in the utilization of alternative data modalities, including skeleton, depth, infrared sequence, point cloud, event stream, audio, acceleration, radar, and WiFi, for HAR \cite{tuo2022long,le2022skeleton,rahmani20163d,jiang2017learning,wang20203dv,ghosh2019spatiotemporal,liang2019audio,zeng2014convolutional,wang2019temporal}. 
However, despite this diversification, RGB video data continues to dominate as the most prevalent data type within the field of action recognition datasets, such as UCF101 \cite{soomro2012ucf101}, HMDB51 \cite{kuehne2011hmdb}, Kinetics \cite{kay2017kinetics}, ActivityNet \cite{kay2017kinetics}, etc \cite{idrees2017thumos,shahroudy2016ntu,goyal2017something,gu2018ava}.
These datasets encompass a wide range of action categories, including basic daily activities, sports activities, gestures and poses, interactive behaviors, entertainment activities, and occupation-specific actions.

Several trends can be observed in the development of HAR video datasets. 
(1) An increasing trend towards larger dataset scales to accommodate a greater number of action instances and diverse scenarios. 
(2) An emphasis on enhancing dataset diversity, including various action classes, backgrounds, environments, and demographics. 
(3) A growing focus on fine-grained annotation, where datasets are annotated at a more detailed and granular level. 
These trends pave the way for tackling more complex and real-world action recognition tasks.

\subsection{Human Action Recognition Methods}
Human action recognition has achieved great improvements in deep learning era. Convolutional neural networks (CNNs) based methods, including 2D CNNs, 3D CNNs and 2D+3D CNNs, effectively extract spatial and temporal features from RGB frames and optical flows. Specifically, 
3D CNNs extend 2D CNN architecture by incorporating the temporal dimension to model spatial feature and temporal dependency simultaneously \cite{qiu2017learning}, and 2D+3D CNNs combine both advantages of them \cite{simonyan2014two,ji20123d,zhang2020few}. 
Transformer-based methods \cite{vaswani2017attention,vaswani2017attention,khan2022transformers} utilize self-attention to capture long-range spatio-temporal relationships, which model the dynamics of human actions by focusing on relevant regions. 
With the improvement of human pose estimation, a promising approach involves first extracting the human body's pose and subsequently performing action recognition \cite{yan2018spatial,sun2019deep,liu2019learning,gong2022meta}.
The skeleton-based human action recognition methods demonstrate more stable performance in certain tasks by mitigating the interference from backgrounds \cite{wang2013learning,vemulapalli2014human,si2019attention}.

\section{Human Habitual Behavior Dataset}

In the field of psychology, HHBs refer to the automated behavioral patterns formed by individuals in specific environments, which have been repeatedly and consistently reinforced to the extent that they can occur spontaneously without conscious involvement \cite{wood2016psychology}.  
Research indicates that the processes of habit formation and maintenance are related to cognition, motivation, and emotion in humans.
HHBs are kinds of external manifestations of human emotions, at the same time, people's emotional and affective states can influence the execution and maintenance of habitual behaviors \cite{neal2011pull,gardner2015review}.
Both of behavioral habits and psychological changes of characters are essential components of video understanding. 
Therefore, identifying HHBs holds significant implications for enhancing the comprehension of video content.

However, the recognition of HHBs has not received widespread attention, not only because such actions themselves are not easily noticed, but more importantly, the collection and annotation of such actions are challenging.
Firstly, in contrast to human actions such as sports activities and interactive behaviors, HHBs are typically inconspicuous throughout the entire video segments. Consequently, these videos are usually not labeled or named based on HHBs, making it challenging to directly locate video clips showcasing HHBs on the internet.
Human motion capture is also a possible method, and there are a lot of HAR datasets recorded manually, like the NTU RGB+D 120 dataset \cite{shahroudy2016ntu}.
However, the HHBs are those tiny actions that arise unconsciously and naturally.
Thus, human recording will greatly reduce the variety of actions and will make the dataset unreal.
Secondly, defining the categories of the dataset is also challenging.
To ensure that the categorized actions can adequately serve the analysis of character traits, habits, and psychological activities, substantial background knowledge in human behavior is required.

\subsection{Dataset Construction}
We build a dataset which contains 30 categories of HHBs, including more than 300,000 frames and 6,899 action instances.
The video clips are cut from the videos on the Internet, including teleplays, dramas, talk shows and daily life videos.
The data collection process mainly consists of three steps: label selection, video clip collection, and category annotation.

\textbf{\emph{Label selection.}} When selecting labels, there are several important factors to consider:
(1) These action categories can reflect the character's emotional changes, behavioral habits, or personality traits.
(2) These actions are all common actions or behaviors in daily life or artistic works.
(3) These actions are rarely present in the existing dataset.
During the label selection process, we extensively refer to behavioral psychology experiments and publicly available videos, including movies, stage plays, and documentaries. 
Throughout this process, we record any actions that meet the above criteria.
As a result, 30 action categories are selected as shown in Fig.~\ref{fig:hhb}.
Most of these actions reflect interactive behaviors between different body parts, such as between hands, hands and face, and so on.

\textbf{\emph{Video clip collection.}} 
In previous datasets, video data is mainly extracted by using the following methods: self-recording, extraction from publicly available films.
We have analyzed earlier that self-recorded video clips, although having higher resolution, richer information, and easier access to video samples, will lead to insufficiently realistic samples for our dataset.
Therefore, we choose to collect video clips containing HHB actions from existing videos on the internet, including TV shows, operas, reaction videos, etc.
Considering that most of the HHB actions are subtle, we have devised a series of strategies to help us retrieve the desired video clips from the vast amount of videos available on the internet.
Firstly, we refer to a series of behavioral psychology evidences to expand our action categories with associated phrases. These phrases are required to frequently co-occur with their corresponding action categories, for example, ``awkwardly touched the nose''.
These expanded textual descriptions can increase the probability of retrieving the desired videos.
Then we search for the corresponding video sources according to the extended phrases.
After cropping the collected videos, we obtain 1947 video clips, each of which contains one or more actions belonging to the aforementioned HHB categories. Most of the videos contain multiple persons, with at least one person performing an action belonging to one of the 30 HHB categories. 
During the data collection, we make efforts to include diverse video sources with varying characters, which ensures that the models trained on our dataset can avoid overfitting to specific backgrounds, clothing, and other unrelated information in the videos.

\begin{figure*}[t!]
    \centering
    \includegraphics[width=\linewidth]{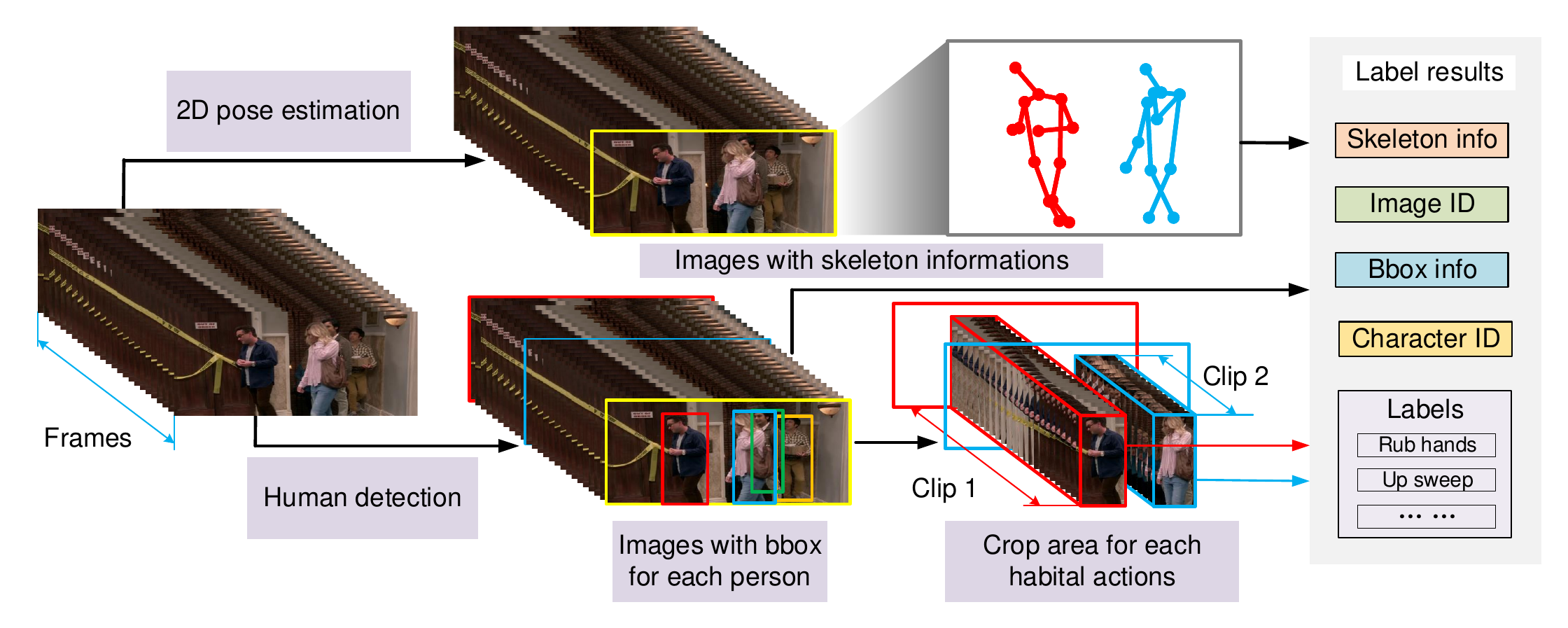}
    \caption{The construction process and data structure of the proposed dataset involve the following steps. Firstly, we utilize the DC-pose model~\protect\cite{liu2019learning} to obtain the bounding box (bbox) and skeleton information for each person in the collected videos. Then we assign action categories to each individual's motions. In the final dataset, we have the positions of each individual, their corresponding skeleton information, and the associated action categories.}
    \label{fig:Construction_process}
\end{figure*}

\textbf{\emph{Category annotation.}}
The form and accuracy of labeling action categories determine the quality of the dataset. Therefore, we not only hope to have precise labeling for the collected videos, but also a more fine-grained labeling format that is convenient to use and can assist downstream tasks. 
For many coarse-grained datasets, only the action categories that appear in a video segment are labeled, which could not describe complex human behavior and interaction information.
To solve this problem, we annotate the action categories for each frame of the collected video segments and label the human bounding box corresponding to each action instance in the video frame.
Many existing human detection algorithms provide convenience for detecting the bounding box of the characters.
In this work, we choose DC-pose model \cite{liu2019learning} to detect the bounding box and the 2D skeleton coordinates of each person frame-by-frame.
For each bounding box, we assign an ID based on its position from left to right in the image, and label its action category. 
The detected 2D human skeletons can be used as another input channel in our proposed HHB recognition model.
The detailed annotation procedure is shown in Fig. \ref{fig:Construction_process}.
Particularly, to reduce prediction errors from DC-pose, we manually correct the obvious inconsistencies of the estimated bounding boxes and skeleton coordinates.

\begin{figure*}[tb]
    \centering
    \includegraphics[width=\linewidth]{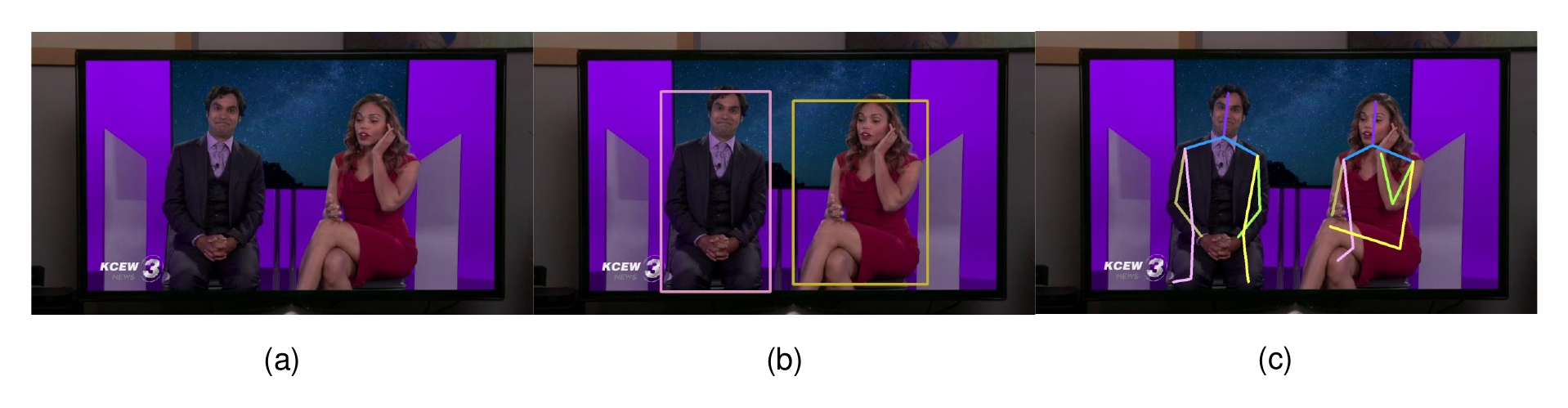}
    \caption{An annotation example of one video clip in the dataset. (a) The original video. (b) The video with human bounding boxes. (c) The video with skeleton information. The two persons are annotated and saved as ``P1\_rubhands'' and ``P2\_crosslegs\&touchear''.}
    \label{fig:s_video}
\end{figure*}

An annotation example is shown in Fig. \ref{fig:s_video}, where there are two characters in this video clip.
The spatial positions and skeleton information of the two characters are shown in Fig. \ref{fig:s_video}(b) and Fig. \ref{fig:s_video}(c), respectively. Based on their positions from left to right, we assign unique identifiers to the two persons, namely ``P1'' and ``P2'', and label their actions as ``rub hands'' and ``cross legs \& touch ear'', respectively. 

\begin{figure*}[tb]
    \centering
    \includegraphics[width=\linewidth]{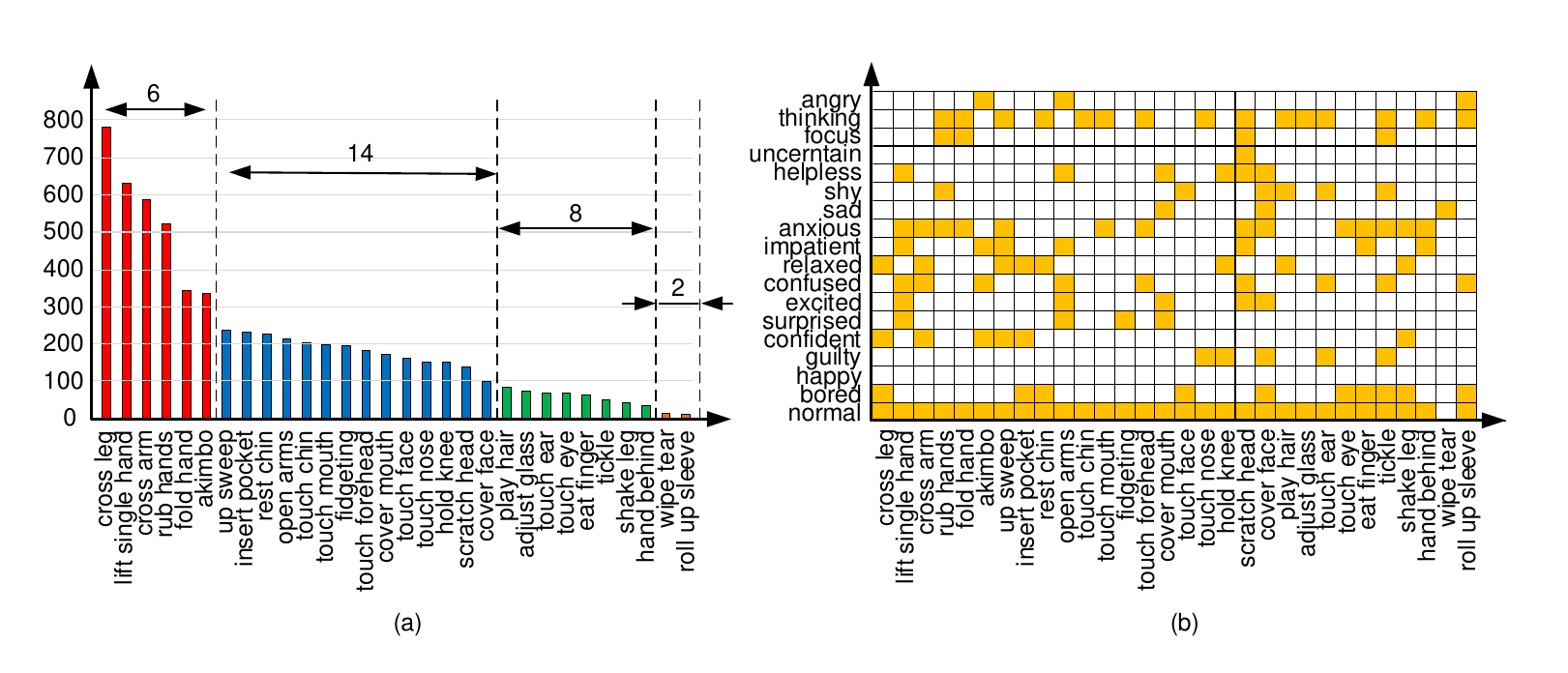}
    \caption{The statistical characteristics of the proposed dataset. The proposed dataset consists of various action categories along with the corresponding sample counts and potential associated emotion categories. (a) Statistical characteristics of the proposed dataset. (b) The emotional attributes reflected by the HHBs.}
    \label{fig:statistical} 
\end{figure*}

\begin{figure*}[t!]
    \centering
    \includegraphics[width=\linewidth]{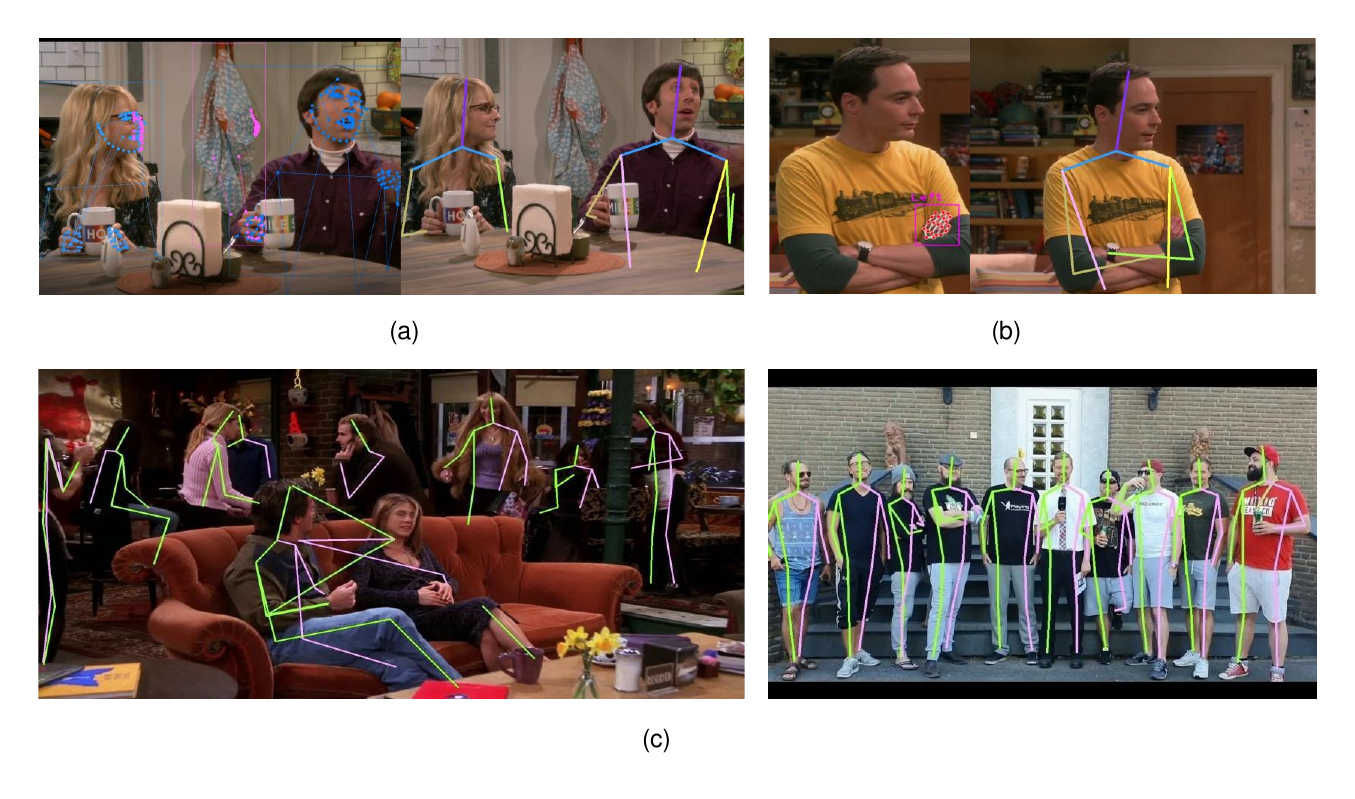}
    \vspace{-18pt}
    \caption{Comparison of skeleton extraction performance across different models. (a) The results obtained through the mmaction2~\protect\cite{2020mmaction2} from the mmlab and DC-pose model \protect\cite{liu2021deep}. There are lots of erroneous predictions in the former results, while the latter have more robust performances. (b) The detection results when using the Mediapipe tool from CVZone and DC-pose model \protect\cite{liu2021deep}. (c) Skeleton extraction results using the DC-pose model \protect\cite{liu2021deep} in a multi-person scene. In this work, we choose the DC-pose to extract the main joints due to its robustness to complex background.
    }
    \label{fig:dcpose}
\end{figure*}

\subsection{Dataset Statistics and Emotional Attributes}
Specifically, this dataset contains 30 action categories and 1,947 video clips, totaling nearly 300,000 frames and 6,899 action instances.
Among the 30 categories, 6 action categories have more than 300 action samples and 20 action categories have more than 100 action samples. The specific number of action samples in each category is shown in Fig. \ref{fig:statistical}(a). 
For each action category, we provide the potential emotional attributes that it may reflect according to the video samples in the dataset and the related psychological literature, such as \cite{lewis1995self,kenny2003action,hertenstein2006touch}, as shown in Fig. \ref{fig:statistical}(b).

These affective attributes are very important for better understanding video content. Detecting the emotional changes of the characters in the video can provide important reference value. 
However, from the perspective of technical implementation, it is very challenging work.
Existing technologies often analyze the changes in a character's facial expressions to understand his or her emotions.
However, the accuracy of this method is greatly affected when the face is obscured.
Our habitual behavior based affective understanding is a good complement to facial expression.
If the algorithm can accurately identify the habitual actions of characters, it will be of great help to deduce the emotional attributes of characters.
For example, when the character appears rubbing hands, scratching the head, touching the nose and some other actions, it can be roughly inferred that the character has a greater probability of being in a negative state such as anxiety or irritability.
And when the character appears to play with hair and other actions, it can be inferred that the task may be in a more relaxed state.
Therefore, the identification of actions in the data set helps to provide important references for the emotional attributes of the task, which are often ignored in the existing emotion recognition tasks.
In addition, our work also helps to provide an important evidence to deduce human characteristics, life experience, psychological state, and so forth.

\section{Human Habitual Behavior Recognition Method} \label{method}

Based on the proposed dataset, we evaluate several mainstream action recognition methods. 
The evaluation process is to retain part of the model parameters obtained by training these models on existing large datasets and finetune them on the proposed dataset.
However, none of the models has outperformed 80\% in terms of accuracy.
The accuracy on the proposed dataset is generally lower than other available datasets.
In order to improve the accuracy of HHB recognition, in this work, we propose a new human action recognition model which combines both skeleton and RGB information. 
The comparison results on this dataset will be described in detail in the experiment section.

\subsection{Two-Channel Feature Extraction}

\begin{figure*}[t!]
    \centering
    \includegraphics[width=\linewidth]{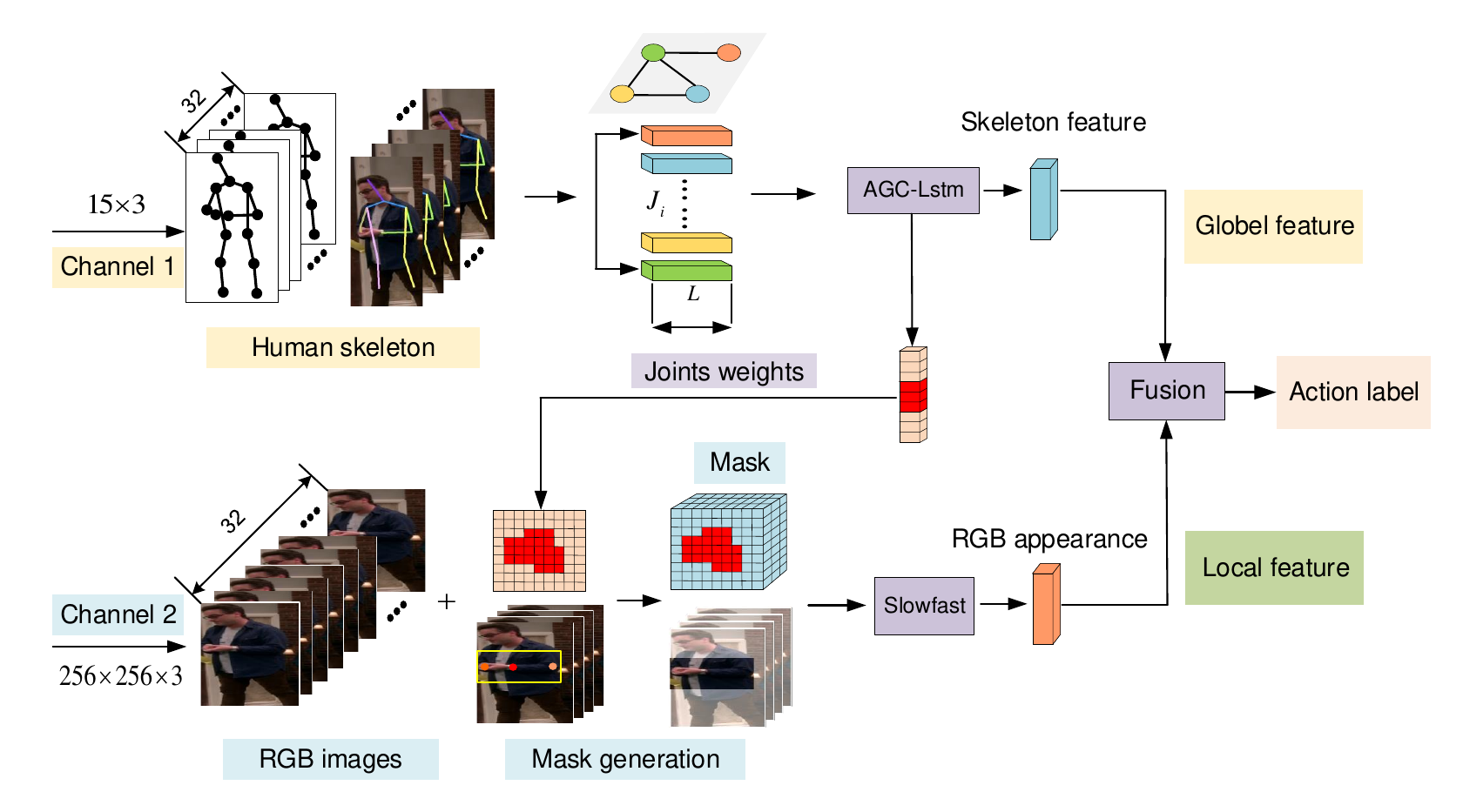}
    \caption{The proposed model framework. There are two channels for extracting skeleton and RGB information, respectively. The skeleton channel focuses on extracting spatio-temporal features and generating a weight vector that represents the correlation between joints and action categories. Based on these weights, an action mask is created to enhance the regions that are highly relevant to the action categories. On the other hand, the RGB channel extracts the detailed  appearance features. Finally, a feature fusion block is constructed to merge the skeleton features and RGB appearance.}
    \label{fig:framework}
\end{figure*}

Regarding the HHB dataset, we propose to extract two-channel feature representations for each frame, namely human skeleton and RGB appearance. Compared with RGB appearance, human skeleton is more effective for HAR, as it only retains useful information related to human posture, greatly reducing the difficulty for the model to focus on useful local actions. When extracting human skeletons from videos, there are several key challenges that must be considered. First, videos on the Internet often have blurry images, due to the fact that they are not usually shot with professional high-speed cameras. 
This poses significant challenges when extracting skeleton information. 
Another issue is that there exists human body occlusion in the collected videos, so the model needs to have high robustness in the presence of occlusion. 
After testing several existing methods, e.g., the mmaction2 ~\cite{2020mmaction2}, Mediapipe ~\cite{lugaresi2019mediapipe} and DC-pose \cite{liu2021deep} shown in Fig. \ref{fig:dcpose}, we finally choose the DC-pose model to preprocess the input video data and obtain the skeleton information of the characters in the video, because DC-pose has high robustness under occluded conditions.
In addition to human skeletons, we also establish a channel for RGB feature extraction of the input image, which can effectively supplement more detailed information, including the missing hand actions often ignored by most of the skeleton extraction methods.

\subsection{Model Framework}

The proposed model framework is shown in Fig. \ref{fig:framework}, which consists of two channels to  process human skeletons and RGB appearance, respectively. In this framework, we add a self-attention module to the human skeleton channel, which enables the model to effectively focus on the local joints of the human body. 
Based on the joint attention weights, we build a spatial mask to select key image regions in each frame, which improves the feature extraction capability of the RGB channel. Finally, a fusion block is established to fuse the features or the scores from the two channels.

\textbf{\emph{Human skeleton channel.}} Recently, there have been many works on skeleton feature extraction. 
We adopt the AGC-LSTM \cite{si2019attention} as the backbone network in our model, which combines graph convolutional networks and LSTM networks to extract both spatial and temporal features. In this channel, a self-attention module is integrated into the feature extraction results, which is not only to enhance the extracted features, but also to determine which joints are more relevant to the actions. 
In this way, we can obtain the weight of each joint with the current action.

Based on the obtained relevance weights and corresponding joint positions, we generate an action mask for each video frame. 
This mask can be considered as a filtering window that retains the related RGB information while filtering out the irrelevant regions with respect to the current action.

\textbf{\emph{RGB appearance channel.}} We adopt the SlowFast model \cite{feichtenhofer2019slowfast} as the backbone network for RGB feature extraction. 
The SlowFast model involves a Slow pathway, operating at low frame rate to capture spatial semantics, and a Fast pathway, operating at high frame rate to capture motion at fine temporal resolution. Therefore, the SlowFast model can effectively adapt to the changes in the speed of actions and has shown good performance on many datasets. We input the visual appearances that has been processed with the mask into the slowfast channel to extract the RGB features, and then fuse the extracted features from the two channels to get the action classification scores.

\textbf{\emph{Channel fusion.}} 
We adopt two fusion strategies: feature fusion and score fusion. 
Feature fusion involves weighting and summing two channel feature vectors before classifying the combined vector. In contrast, score fusion directly weights and sums the classification probabilities from the two channels.

\subsection{Architecture Details}
\textbf{\emph{Skeleton feature representation.}} 
The AGC-LSTM adopts GCN model to build the representation of human joints. 
In this work, the human joints representation, which includes 15 main joints, are extracted by DC-pose, as shown in Fig.~\ref{fig:framework}.
Based on the coordinate information of skeleton joints in the videos, the input data of the skeleton input channel is shaped into $(b, L, m, n)$, where $b$ denotes the batch size, $L$ the length of the input sequence, $m$ the number of the joints, $n$ the dimension of joint coordinates.
In this work, every video clip are preprocessed as tensors with a length of 32, representing the information of the input data in time sequence.
The $n$ is set as 3 in this work. Among these 3 dimensions, the first two dimensions are the $x$ and $y$ coordinates of every joint, and the third dimension represents the confidence coefficient of the coordinates.
Finally, the output feature size of the skeleton input channel is $(b,i)$, which $b$ is the batch size, and $i$ is the total number of the action categories.

\textbf{\emph{RGB feature representation.}} The RGB feature represents the RGB information of the input video clips. 
The size of the RGB feature is shaped into $(b, c, L, w, h)$, where $b$ denotes the batch size, $c$ the number of RGB channels, $L$ the length of the input sequence, $w$ and $h$ the width and height of every images.
The output feature size of the RGB input channel is the same as the skeleton input channel.

\textbf{\emph{Action mask.}}
The mask is built to enhance the spatial information that is highly related to the habitual actions.
Since these habitual actions usually appear at the local parts of the persons, the feature enhancement is critical so that the model can pay more attention to the local features.
We adopt a self-attention module to extract the correlation between the joints and the action categories, and generate a mask to enhance the spatial information in each video.
Specifically, as shown in Fig.~\ref{fig:framework},
the correlation is shaped as a one-dimensional vector with a length of $m$ (i.e., the joint weight vector in Fig.~\ref{fig:framework}).
Next we extract the coordinates of the top three related joints and set the coordinates as the centers of child masks.
The sizes of these masks are set according to the sizes of each person's bounding box with a specific scale ratio.
In the data preprocessing stage, each person's video clip will be reshaped into $256\times 256 \times t$ in each RGB channel according to the bounding box provided by DC-pose. $t$ is the time length of the sampled video.
Therefore, the size of each mask is set as $128 \times 128$, which is half of the length and width of each person's bounding box. 
By overlaying all the child masks, we can obtain the final mask (i.e., the mask in Fig.~\ref{fig:framework}).
If the pixel is located in the mask, the value is unaltered; otherwise, it will be reduced by multiplying a much smaller value $p$.

The pixel values after masking are updated as follows:
\begin{equation}
{v_{x,y}} = \left\{ {\begin{array}{*{20}{c}}
  {{v_{x,y}},(x,y) \in \Omega} \\ 
  {p\times {v_{x,y}},(x,y) \notin \Omega} 
\end{array}} \right.
\label{eq1}\end{equation}

Where $v_{x,y}$ is the value of the pixel $(x,y)$, $\Omega$ is the scope of the mask. The coordinate $(x,y)$ in mask $\Omega$ should satisfy the constraint:
\begin{equation}
\{ (x,y) \in \Omega | \exists (x_i, y_i) \in \mathcal{C}, (|x - {x_i}| < \frac{l_x}{2}) \wedge (|y - {y_i}| < \frac{l_y}{2})
\}\label{eq2}
\end{equation}

Where $l_x$,$l_y$ are the width and height of the mask, $\mathcal{C}=\{(x_i,y_i)\}_{i=0}^2$ is the set of three mask centers.

\begin{figure*}[t!]
    \includegraphics[width=\linewidth]{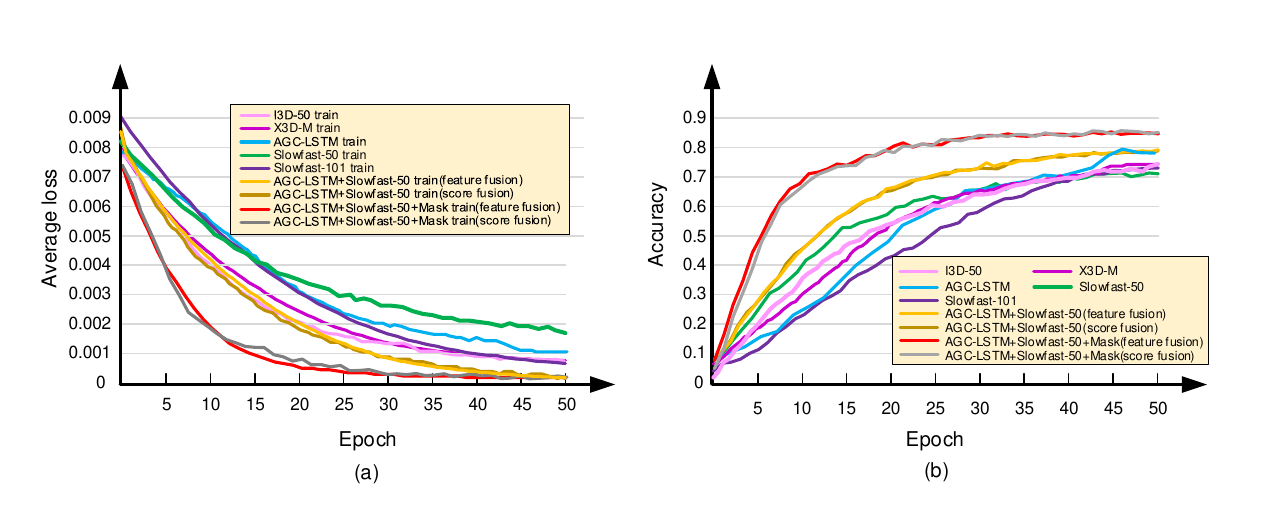}
    \caption{(a) Loss curves and (b) accuracy curves of our models and several baselines. Our model with action mask under the either feature fusion or score fusion all achieves the best performance.}
    \label{fig:lineplot}
\end{figure*}

\section{Experiments} \label{experiment}

\subsection{Implementation Details}
To verify the effectiveness of the proposed model in our HHB dataset, we conduct a series of comparison experiments.
In these experiments, all the video clips are divided into training, test sets, and validation sets in a ratio of 0.7, 0.2 and 0.1.
In addition, we have tried to avoid video clips from the same video source being distributed in both the training set and the test set.
Particularly, the training and test sets are divided in each category to prevent uneven training data across certain categories.
During the test, each video clip is shaped into the size of $(18,32,15,3)$ for the skeleton input channel and the size of $(18,3,32,256,256)$ for the RGB input channel.
More specifically, for the slowfast model within the RGB input channel, the $(18,3,32,256,256)$ tensor is reshaped into $(18,3,8,256,256)$ for the fast branch.
For the action mask generation, the width and height of each child mask are set as $(64,64)$, and the mask parameter $p$ is set as 0.3 to weaken the less related information.
Finally, the batch size is set as 18.

\subsection{Experimental results}

We compare the performance of our model with several existing methods on the proposed dataset.
Specifically, we choose the AGC-LSTM \cite{si2019attention}, Slowfast-50 \cite{feichtenhofer2019slowfast}, Slowfast-101 \cite{feichtenhofer2019slowfast}, I3D-50 \cite{carreira2017quo}, and X3D-M \cite{feichtenhofer2020x3d} as the baselines for the proposed dataset.
cThe AGC-LSTM uses the skeleton information as input and is trained from scratch without pretraining, while the others directly use RGB information and are pretrained on existing datasets, e.g., Slowfast-50, Slowfast-101, I3D-50, X3D-M are pretrained on Kinetics\cite{kay2017kinetics}. Similarly, our model adopts a Slowfast-50 model pretrained on Kinetics in the RGB channel and a no-pretrain AGC-LSTM network in the skeleton channel. 
The accuracy is obtained by dividing the number of correctly classified samples by the total number of the samples in the test dataset, and the loss is calculated with a cross entropy function. The accuracy curves and loss curves of our model and the comparison methods are shown in Fig. \ref{fig:lineplot}, and the experimental results are shown in Table~\ref{tab:result}. 
We can see that the variants of our model all surpass 80\% accuracy and perform much better than the comparison methods. Particularly, the action mask extracted from skeleton joint attention weights plays an important role in performance improvement (more than 4\%). It demonstrates that our model can capture the most related local parts of human habitual behaviors.

\begin{table}[!t]
  \caption{Recognition accuracies of the proposed model and the selected baselines on the HHB dataset. The AGC-LSTM model has not been pre-trained on any dataset, while the RGB based models are pre-trained on the Kinetics or EPIC Kitchens dataset. The AGC-LSTM model has the best performance among the single-channel models. The two-stream models generally have better performances than the single-channel models. Our proposed model with action mask under the either feature fusion or score fusion all achieves the best performance among the two-stream models.}
  \label{tab:result}
  \centering
  
  \begin{tabular}{llll}   
    \toprule
    Model            &  Pretrain Dataset   & \multicolumn{1}{c}{Accuracy} \\
    \midrule
    AGC-LSTM             &  no-pretrain        &  \multicolumn{1}{c}{78.5\%}    \\
    Slowfast-50      &  Kinetics           & \multicolumn{1}{c}{72.4\%}    \\
    Slowfast-101    &  Kinetics           & \multicolumn{1}{c}{72.7\%}    \\
    I3D-50 &      Kinetics           & \multicolumn{1}{c}{73.1\%}    \\
    X3D-M &        Kinetics           & \multicolumn{1}{c}{72.8\%}    \\
    \midrule
    AGC-LSTM+Slowfast-50         & Kinetics  & \multicolumn{1}{c}{80.3\%}  \\
    (feature fusion)\\
    AGC-LSTM+Slowfast-50             & Kinetics  & \multicolumn{1}{c}{80.1\%}  \\
    (score fusion)\\
    AGC-LSTM+Slowfast-50  &     Kinetics  & \multicolumn{1}{c}{84.5\%}  \\
    +Mask (feature fusion)\\
    AGC-LSTM+Slowfast-50     & Kinetics  & \multicolumn{1}{c}{84.4\%}  \\
    +Mask (score fusion)\\
    \bottomrule
  \end{tabular}
\end{table}

\textbf{\emph{Comparison with large pre-trained models.}}
We compare our model with a large video model InternVideo \cite{wang2022internvideo} and a large image-text model CLIP \cite{radford2021learning} on the proposed HHB dataset. In our experiments, InternVideo uses the provided B/16 and ViT-B-16 models while CLIP adopts the pretrained ViT-B/32 model as the backbone networks. Since CLIP is mainly used to handle static images, we randomly capture a video frame as the input of the model. During action recognition with these two models, we extract features from pre-trained backbone networks and train linear classifiers to the 30 HHB categories. The recognition accuracies of both InternVideo and CLIP are presented in Table \ref{tab:internvideo}, which are much worse than those results mentioned in Table \ref{tab:result}.  
The significant performance gaps indicate that large pre-trained video models struggle to extract effective features for human habitual behavior (HHB) videos. This suggests that the training corpora for these models likely lack sufficient HHB video content. Therefore, our proposed HHB dataset is a crucial addition to human action datasets.

\begin{table}[!t]
  \caption{Comparison results of two large pre-trained models (InternVideo\cite{wang2022internvideo} and CLIP\cite{radford2021learning}) .}
  \label{tab:internvideo}
  \centering
  
  \begin{tabular}{lll}   
    \toprule
    Model           &  InternVideo\cite{wang2022internvideo}   & CLIP\cite{radford2021learning}   \\

    \midrule
    Accuracy &  27.9\%   & 20.1\%    \\ 
        
    \bottomrule
  \end{tabular}
\end{table}

\textbf{\emph{Human skeleton vs RGB images.}} As shown in Fig.~\ref{fig:lineplot} and Table.~\ref{tab:result}, although the details about hands and fingers are not included in the human skeleton data, the AGC-LSTM still performs much better than the other baseline models that adopt RGB images as input.
One possible reason for this phenomenon is that human skeletons represent the core structure of motion and are less prone to noise and distractions.

\textbf{\emph{AGC-LSTM and Slowfast.}} 
In order to verify whether the combination of skeleton and RGB features can improve the model performance, we combine AGC-LSTM and Slowfast together by using feature fusion and score fusion, respectively.
The outputs of the two fused models are set with the same size as $(b,i)$.
The results show that there is about 2\% improvement than the best single-channel model regarding the accuracy (80.3\% vs 78.5\%), which demonstrates that human skeleton and RGB appearance can complement each other to improve model performance.

\textbf{\emph{AGC-LSTM + Slowfast vs AGC-LSTM + Slowfast + Mask.}} 
Considering the advantages of combining human skeleton and RGB image, we propose a more effective strategy to integrate these two modalities by using action masks. Therefore, it is necessary to compare the model performance in cases with and without action masks to evaluate the effectiveness of this approach.
As we have anticipated, the results indicate that applying masks improves the model performance by approximately 4\% compared to direct combination without masks.
This means that paying closer attention to the local parts selected by the self-attention module plays a significant role in identifying these subtle habitual actions.

\begin{figure*}[!t]
    \centering
    \includegraphics[width=\linewidth]{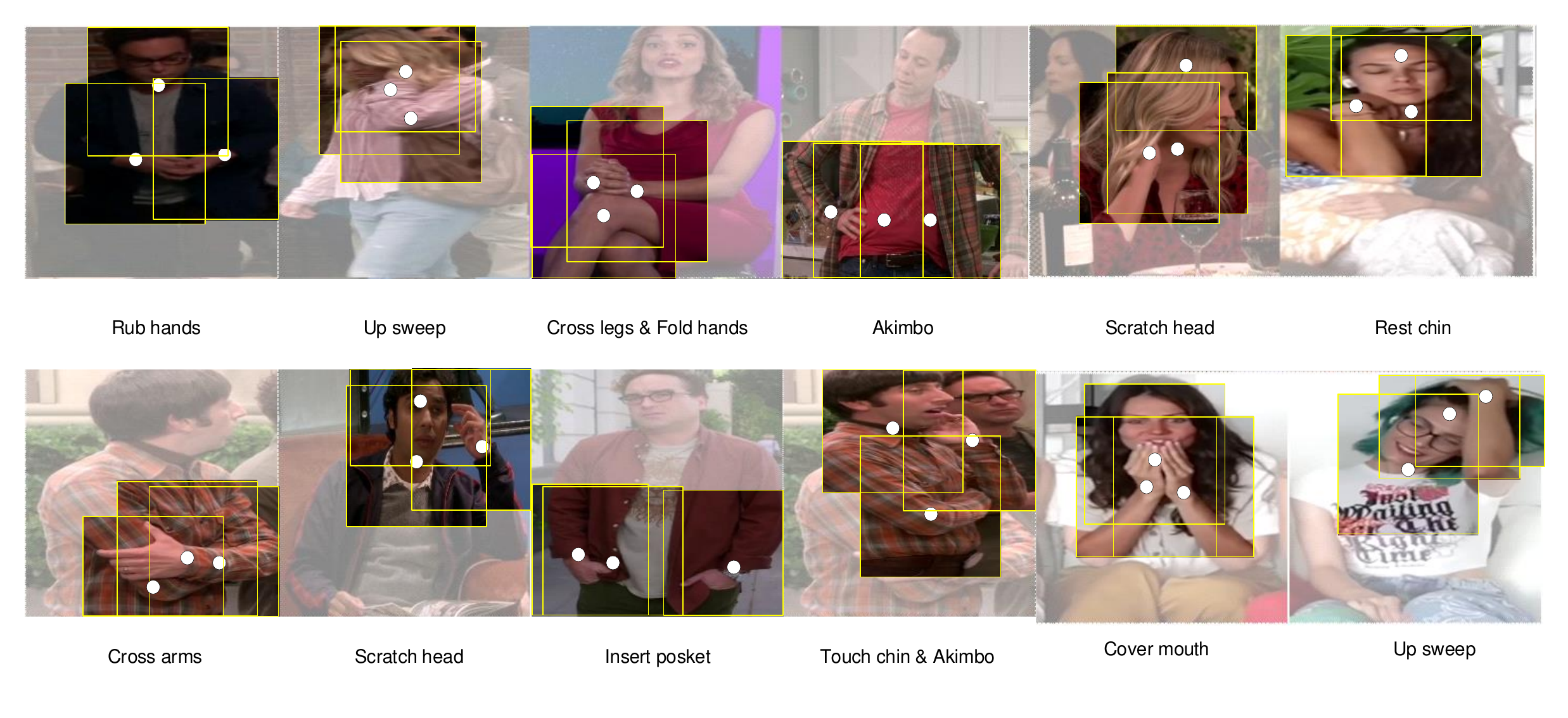}
    \caption{Action mask visualization. The three highlighted points in each image represent the positions corresponding to joints with more significant weights. Three child masks were created with these points as their center points for each video. It can be seen that the most-related regions to human habitual behavior are selected. }
    \label{fig:mask}
\end{figure*}

\textbf{\emph{Action Mask Visualization.}} 
In order to demonstrate the masks more intuitively, we choose several video clips and show the enhanced single frame in Fig.~\ref{fig:mask}.
The highlighted points in each frame indicate the positions of the selected joints, which also serve as the centers of the child masks. It can be seen that these masks significantly enhance the RGB features of each frame in the video clips.

\section{Conclusion}

In this paper, we introduce a novel video dataset showcasing various human habitual behaviors (HHBs) and a carefully designed action recognition model. To the best of our knowledge, this is the first HAR dataset focusing specifically on human habitual behaviors. The action categories in this dataset reveal implicit information about individuals' personality traits, lifestyle habits, and psychological changes. Detecting these actions is crucial for a deeper understanding of video content. Moreover, our proposed two-stream model offers an effective solution for recognizing subtle actions. Experimental results demonstrate that our model outperforms existing methods in recognizing human habitual behaviors.

\bibliographystyle{abbrv}



\end{document}